\documentclass[journal]{IEEEtran}
\usepackage[utf8]{inputenc}
\usepackage{graphicx}
\usepackage{booktabs}
\usepackage{url}
\usepackage{cite}
\usepackage{xcolor}
\usepackage{amsmath}
\usepackage{amssymb}

\newcommand{\SE}{\mathrm{SE}(3)}

\usepackage{hyperref}
\usepackage{orcidlink}

\hypersetup{hidelinks}

\begin{document}

\title{J-LAW: Joint Localization and Action-Conditioned World Modeling via Coupled Latent Factor Graphs}

\author{\IEEEauthorblockN{Guanqun Cao}$^{\orcidlink{0000-0002-7014-219X}}$,~\IEEEmembership{IEEE Member}, \IEEEauthorblockN{Liang Chen}$^{\orcidlink{0000-0002-7083-6001}}$,~\IEEEmembership{IEEE Member}
\thanks{G. Cao is with Geely Technology Europe, Pumpgatan 1, 417 55 Gothenburg Sweden (e-mail: Guanqun.Cao@ieee.org).}
\thanks{L. Chen is with the State Key Laboratory of Information Engineering in Surveying, Mapping and Remote Sensing (LIESMARS) , Wuhan University, China, (e-mail: l.chen@whu.edu.cn).}}
\graphicspath{{./figures/}}
\maketitle

\begin{abstract}
Classical simultaneous localization and mapping (SLAM) estimates metric poses and a geometric map but does not provide an action-conditioned predictive state. Action-conditioned world models learn compact latent dynamics but ignore global metric consistency and accumulate drift under open-loop rollout. We introduce J-LAW (Joint Localization and Action-Conditioned World Modeling), a unified factor-graph formulation that connects metric pose variables, predictive latent states, and persistent latent landmarks in this letter.
J-LAW represents each image as a compact predictive state and combines it with pose or motion measurements through a separately learned mapping.
Its maximum a posteriori (MAP) factor graph enforces consistency between these complementary sources of information over time. 
Experiments on PushT and WildGS show that J-LAW's factor-graph representation can improve long-horizon latent consistency and recover more reliable predictive states under partial observations, forming a foundation for future integrated localization and planning systems.
\end{abstract}


%==================================================================
\section{Introduction}
%==================================================================

Two essentials are needed for a mobile agent to operate in the real world at once: \emph{where it is} (localization) and \emph{what will happen next} (world modeling). These are traditionally solved by separate systems.
Classical SLAM estimates a trajectory and a geometric representation of the environment from sensory observations. Its key strength is global geometric consistency: odometry and loop-closure constraints can be jointly optimized to reduce accumulated localization error. However, the mapping by a conventional SLAM is not a transition model, and it does not explicitly predict how future states or task-relevant latent variables change after a control input.

Action-conditioned world models address the complementary prediction problem. They encode observations into a compact state and learn a transition model that predicts a future state from the current state and an action. Such models can support short-horizon prediction, but their latent trajectories are commonly evaluated in open loop and are not constrained by metric pose-graph structure or revisits. Our goal is therefore to jointly solve predictive latent dynamics and geometric constraints together.

We propose J-LAW, a coupled factor graph (Fig.~\ref{fig:teaser}) that jointly estimates metric poses $X=\{x_t\}$, latent world states $Z=\{z_t\}$, and latent landmark embeddings $M=\{m_k\}$.
It enforces learned transition, loop, and landmark factors with constraints that connect pose or motion measurements to visual predictive states for offline predictive-state inference.
We evaluate this formulation on PushT and WildGS datasets. Our main contributions are:
\vspace{-1mm}
\begin{figure}[t]
\centering
\includegraphics[width=\columnwidth]{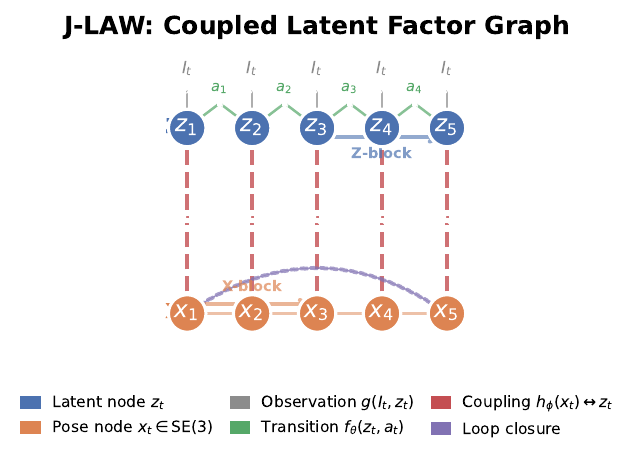}
\caption{J-LAW coupled factor graph. In the WildGS experiment, latent nodes $z_t$ (blue, top) and pose nodes $x_t \in \SE$ (orange, bottom) are linked by a learned pose-to-latent coupling factor $h_\phi$ (red dashed). Observation factors $g$, transition factors $f_\theta$, and loop closure (purple) constrain the corresponding variables. Alternating block coordinate descent optimizes the $X$-block and $Z$-block in turn.}
\label{fig:teaser}
\end{figure}

\begin{itemize}
    \item We propose a maximum a posteriori (MAP) factor-graph formulation that connects geometric state estimation and action-conditioned latent prediction through learned measurement-to-latent coupling factors.
    \item We use frozen visual latent embeddings with action-conditioned transition factors to predict future states, while revisit and landmark factors refine predictive latent trajectories.
    \item Our evaluation on PushT and WildGS shows improvement of long-horizon latent consistency relative to open-loop rollout and predictive-state recovery under missing observations, while characterizing coupled latent--pose inference on WildGS.
\end{itemize}

%==================================================================
\section{Related Work}
%==================================================================

The goal of SLAM is to estimate the pose of a mobile agent while constructing a map of its surrounding environment. SLAM has supported a wide range of vision and robotics applications by estimating trajectories and geometric maps from sensor observations~\cite{Thrun2008}. Factor-graph smoothing provides an efficient framework for incremental nonlinear optimization of these geometric variables~\cite{Dellaert2017,cCatal2021}. However, traditional SLAM formulations do not explicitly connect geometric state estimation to learned predictive latent representations or action-conditioned dynamics. This decoupling leaves a gap between abstract behavioral planning and precise spatial grounding.
Lee et al.~\cite{Lee2013} optimize metric poses and geometric maps with expectation-maximization but produce no predictive, action-conditioned model.
Notably, predictive spatial mapping~\cite{Gornet2024} was established that visual next-state prediction task objectives force neural latent spaces to implicitly organize into metric-accurate cognitive maps. It provides strong empirical validation that a system optimized for predicting sensory consequences implicitly resolves global environment geometry, embedding the map directly in the predictive coding.

Action-conditioned world models / JEPAs are effective ways of predicting the outcomes of actions through learning from the environment~\cite{Maes2026}. Recent work in JEPAs concentrates on the most relevant features used for future states and produces a compact, low-dimensional latent space and temporal model by predicting the latent representation of future observations.
For example, LeWorldModel~\cite{Maes2026} learns stable latent dynamics from pixels, and V-JEPA series~\cite{Assran2025, Mur-Labadia2026} extends joint-embedding prediction to video representation learning.
Moreover, SkyJEPA~\cite{Rao2026} adopts JEPA backbones combined with kinematic probers to achieve accurate, long-horizon predictions for real-time closed-loop quadrotor control, but they focus solely on high-frequency local flight dynamics.
These methods yield predictive latent states but lack global consistency and loop closure.
In contrast, J-LAW adds a factor‑graph inference layer to enforce consistency without retraining the predictor.
% In contrast, J-LAW introduces latent landmarks that to achieve both purposes: they serve simultaneously as a low-dimensional, predictive planning substrate for System 2 rollouts and as a geometrically consistent topological map for global state estimation.

%==================================================================
\section{J-LAW Formulation}
%==================================================================

\subsection{Coupled State}

J-LAW maintains a coupled state over a trajectory of length $T$:
\begin{equation}
    \mathcal{S} = \big\{ X_{1:T},\; Z_{1:T},\; M_{1:K} \big\},
\end{equation}
where $X_{1:T}=\{x_1,\dots,x_T\}$ are metric poses,
$Z_{1:T}=\{z_1,\dots,z_T\}$ are predictive latent states, and
$M_{1:K}=\{m_1,\dots,m_K\}$ are persistent latent landmark embeddings.

A frozen visual encoder provides observation-derived latent embeddings:
\begin{equation}
    \bar z_t = E_\theta(o_t),
\end{equation}
where $o_t$ is the visual observation at time $t$ and $\bar z_t$ is the
encoder target. When pose or motion measurements are available, J-LAW
denotes them by $s_t$ and uses a separately trained mapping $h_\phi$ to
predict the corresponding latent state:
\begin{equation}
    h_\phi(s_t) \approx z_t.
\end{equation}
For WildGS, $s_t$ is a metric pose; for the PushT missing-observation
experiment, $s_t$ is recorded proprioception (agent position and velocity). The factor graph combines this
measurement-derived prediction with visual latent observations and learned
action-conditioned dynamics during inference.

\subsection{Joint MAP Objective}

We estimate the coupled state by solving a maximum a posteriori problem:
{\small
\begin{align}
\label{eq:jlaw}
\{X^\star, Z^\star, M^\star\} = \arg\min_{X,Z,M} \; \Big\{ \Phi_{\mathrm{obs}} + \lambda_p \Phi_{\mathrm{pred}} + \lambda_g \Phi_{\mathrm{mot}} \nonumber \\
+ \lambda_c \Phi_{\mathrm{couple}} + \lambda_l \Phi_{\mathrm{loop}} + \lambda_m \Phi_{\mathrm{lmk}} \Big\} ,
\end{align}}
with the six factor terms:
%The full objective describes the pose--latent instantiation used for
%WildGS; the PushT experiments use the corresponding latent-only subgraph
%and, during outages, instantiate $s_t$ with recorded proprioception. The
%factor terms are:

\paragraph{Latent observation factor}
Anchors each latent state to the frozen visual encoder embedding, with a
robust kernel:
\begin{equation}
    \Phi_{\mathrm{obs}} =
    \sum_{t=1}^{T}
    \rho\!\left(
    \big\| z_t - E_\theta(o_t) \big\|^2_{\Sigma_o^{-1}}
    \right).
\end{equation}

\paragraph{Action-conditioned latent prediction}
Enforces JEPA-style predictive consistency between consecutive latents:
\begin{equation}
    \Phi_{\mathrm{pred}} = \sum_{t=1}^{T-1} \big\| z_{t+1} - P_\theta(z_t, a_t) \big\|^2_{\Sigma_p^{-1}}.
\end{equation}

\paragraph{Metric motion model}
Standard SLAM odometry factor on poses (IMU preintegration or kinematic):
\begin{equation}
    \Phi_{\mathrm{mot}} = \sum_{t=1}^{T-1} \big\| x_{t+1} \boxminus g(x_t, a_t) \big\|^2_{\Sigma_g^{-1}},
\end{equation}
where $g(x_t,a_t)$ denotes a metric motion model and $\boxminus$ is the
$\SE$ between operation.

%\paragraph{Metric pose-loop factor}
%For the supplied pose-loop pairs used in the WildGS diagnostic, the metric
%trajectory is also constrained by
%\begin{equation}
%    \Phi_{\mathrm{loop}}^X =
%    \sum_{(i,j) \in \mathcal{L}} \big\|x_i-x_j\big\|^2_{\Sigma_x^{-1}}.
%\end{equation}

\paragraph{Measurement-to-latent coupling factor}
A separately trained mapping predicts the latent state from pose or motion
measurements:
\begin{equation}
    \Phi_{\mathrm{couple}} =
    \sum_{t=1}^{T}
    \big\| z_t - h_\phi(s_t) \big\|^2_{\Sigma_c^{-1}}.
\end{equation}
For WildGS, $s_t=x_t$ is the metric pose, so this term couples pose and
latent inference. For PushT, $s_t$ is recorded proprioception and the term
provides complementary information during visual-latent outages. In either
case, it regularizes the predictive latent state using a separately trained
mapping from measurements to latent states.

\paragraph{Latent loop closure}
For a proposed revisit pair, latent states are pulled toward consistency,
with a robust kernel to resist false loops:
\begin{equation}
    \Phi_{\mathrm{loop}} = \sum_{(i,j) \in \mathcal{L}} \rho\!\left( \big\| z_i - z_j \big\|^2_{\Sigma_l^{-1}} \right),
\end{equation}
where $\mathcal{L}$ is the set of latent loop-closure edges.
\paragraph{Latent landmark observation}
Associates latent states with persistent latent landmarks:
\begin{equation}
    \Phi_{\mathrm{lmk}} = \sum_{(t,k) \in \mathcal{O}} \big\| z_t - m_k \big\|^2_{\Sigma_m^{-1}}.
\end{equation}

In the general formulation, pose variables are constrained by geometric
motion and revisit constraints when available, while latent variables are
constrained by visual encoder embeddings and the learned action-conditioned
transition model. The measurement-to-latent factor incorporates additional
measurement evidence during inference, rather than applying a latent smoother
after an open-loop rollout. 
%In WildGS, alternating updates optimize pose and latent
%variables through this compatibility term; in PushT, the evaluated graph
%uses latent-only correction or proprioception-to-latent coupling with missing 
%visual observations.
%\vspace{-5mm}
\subsection{Alternating Block Coordinate Descent}

For the WildGS pose--latent experiment, direct joint optimization over
$X$ and $Z$ is ill-conditioned: when both variables are free, the coupling
factor admits solutions in which $z_t$ and $h_\phi(x_t)$ move toward each
other without improving either estimate. Empirically, joint L-BFGS yields
latent RMSE worse than the latent-only baseline
(Section~\ref{sec:expts}).

We therefore use alternating block coordinate descent in this
experiment: a pose block $\mathcal{B}_X=\{X\}$ and a latent block
$\mathcal{B}_Z=\{Z,M\}$ are optimized in turn while the other block is
fixed.

\paragraph{Pose block (fix $Z$, optimize $X$)}
With latents frozen, the coupling factor becomes a proper regularizer pulling each pose toward the location whose decoded latent matches the current $z_t$:
\begin{equation}
    X^{(k+1)} = \arg\min_X \; \lambda_c \Phi_{\mathrm{couple}}(X; Z^{(k)}) + \lambda_g \Phi_{\mathrm{mot}} + \lambda_l \Phi_{\mathrm{loop}}^X,
\end{equation}
where $\Phi_{\mathrm{loop}}^X$ denotes metric pose-graph loop closure.

\paragraph{Latent block (fix $X$, optimize $Z, M$)}
With poses frozen, the latent sub-problem recovers the standard FG-JEPA objective augmented by the coupling factor:
\begin{align}
    \{Z^{(k+1)}, M^{(k+1)}\} = & \arg\min_{Z,M} \; \Phi_{\mathrm{obs}} \nonumber\\ 
    & + \lambda_p \Phi_{\mathrm{pred}} + \lambda_c \Phi_{\mathrm{couple}}(Z; X^{(k+1)}) \nonumber \\ 
    & + \lambda_l \Phi_{\mathrm{loop}} + \lambda_m \Phi_{\mathrm{lmk}}.
\end{align}
Each block is solved with L-BFGS and strong Wolfe line search. We set $K{=}5$--$8$ iterations. This decomposition ensures that in each block the coupling factor acts as a one-directional regularizer, avoiding the degenerate joint solution while still transferring information bidirectionally across rounds.

\subsection{Training and Inference}

\textbf{Phase 1 (Pretraining).} Train or adopt a visual encoder
$E_\theta$ and action-conditioned latent predictor $P_\theta$. The encoder
receives visual observations only.

\textbf{Phase 2 (Grounding).} Freeze $E_\theta$ and train $h_\phi$ to
predict visual encoder embeddings from the available pose or motion
measurements by minimizing $\|E_\theta(o_t)-h_\phi(s_t)\|^2$. For WildGS,
the measurements are motion-capture poses; for PushT outage imputation,
they are recorded proprioception.

\textbf{Phase 3 (Inference).} At test time, freeze $E_\theta$,
$P_\theta$, and $h_\phi$. WildGS applies alternating pose--latent
optimization, while PushT applies latent-graph correction and, during
outages, optional proprioception--latent coupling.

%==================================================================
\section{Experiments}
\label{sec:expts}
%==================================================================

We evaluate graph-based latent correction and masked-latent state estimation on PushT by latent states from a trained LeWM checkpoint~\cite{Chi2025, Maes2026}, and evaluate the coupled pose--latent method on WildGS dynamic-SLAM sequences with motion-capture poses~\cite{Zheng2025}. The experiments address the following questions: whether factor graph correction can reduce open-loop latent error, whether revisit constraints improve relative consistency while preserving the pointwise correction, whether learned loops and persistent latent landmarks contribute under masking by hidden visual or latent observations, and whether motion information helps state estimation while visual information is missing.

\subsection{State Prediction and Recovery on PushT}
For PushT, we use recorded HDF5 trajectories from the simulated
manipulation benchmark and a frozen trained LeWM checkpoint to construct
latent factor graphs over short trajectory windows. We first compare
one-step latent prediction using observed latent inputs with open-loop
rollout, in which predictions are recursively fed back. We then optimize
latent correction variables in a first-order factor graph and select factor
weights that improve pointwise latent RMSE without excessively constraining
the trajectory.
Finally, we evaluate relative-loop factors and relinearized transition
targets over multiple outer iterations. Here, ground-truth loops are known revisit pairs derived from task state. They are used only as diagnostic graph constraints, not as model inputs.
Representative configurations are reported in
Table~\ref{tab:push}, while the separate latent-only WildGS diagnostic is
reported in Table~\ref{tab:wildgs-latent}.

Two held-out tests of recovering missing latent states are reported in Table~\ref{tab:pusht-imputation}. Firstly, we hide selected visual observations
within short trajectory segments. The model can use its learned dynamics,
earlier latent observations, and, when included, learned revisits and
persistent landmarks to fill in the missing states. The state-derived loops and landmarks serve as ground-truth references.
In the second test, visual observations are removed for a continuous later
part of each segment, while only actions and agent position and velocity remain
available. Here we study only whether these motion measurements help recover
the missing latent states, without revisit or landmark factors. Results
are averaged over $2,903$ windows from held-out trajectories.

\begin{table}[t]
\centering
\caption{PushT latent-rollout correction with supplied GT-state revisit pairs
(diagnostic). End. error is final latent-state error; rel. reduction is the
decrease in relative-revisit residual. $^\dagger w_{\mathrm{loop}}=20$,
$c=0.5$; $^\ddagger w_{\mathrm{loop}}=0.1$, $c=0.1$.
Here, $w_{\mathrm{loop}}$ is the loop-factor weight and $c$ is the
Cauchy-loss scale.}
\label{tab:push}
\footnotesize
\begin{tabular}{lccc}
\toprule
Config. & Latent RMSE & End. error & Rel. reduction \\
\midrule
Open-loop rollout & 1.401 & 21.764 & -- \\
FG, no loop factor & 0.033 & 0.445 & -- \\
FG, GT-state revisits$^\dagger$ & 0.031 & 0.443 & 99.93\% \\
FG, GT-state revisits$^\ddagger$ & 0.032 & 0.438 & 97.91\% \\
\bottomrule
\end{tabular}
\end{table}

\begin{table}[t]
\centering
\caption{WildGS latent-only graph correction (mean over four test scenes).}
\label{tab:wildgs-latent}
\footnotesize
\begin{tabular}{lcc}
\toprule
Method & Latent RMSE & Endpoint error \\
\midrule
Open-loop rollout & 0.31 & 2.52 \\
Latent-only factor graph & 0.09 & 0.95 \\
\bottomrule
\end{tabular}
\end{table}

\begin{table}[t]
\centering
\caption{PushT masked-latent state-estimation and visual-outage imputation. The metrics in the upper half are held-out masked latent RMSE over $2,903$ windows; ground-truth-derived rows are non-deployable upper-bound diagnostics. The metrics in the second half are held-out outage latent RMSE; $w_{\mathrm{couple}}=3$ was selected on validation only by its paired advantage over the cross-episode permutation control.}
\label{tab:pusht-imputation}
\footnotesize
\begin{tabular}{lcc}
\toprule
Setting & Method & Latent RMSE \\
\midrule
Masked & No loop & 1.411 \\
 & Learned loops & 1.407 \\
 & Learned loops + landmarks & 1.395 \\
 & Ground-truth loops + landmarks & 1.363 \\
\midrule
Outage & No coupling & 1.551 \\
 & Direct proprio predictor & 0.945 \\
 & Paired coupling & 0.951 \\
 & Permuted coupling control & 0.965 \\
\bottomrule
\end{tabular}\\[-1mm]
%\raggedright\scriptsize $^\ast$State-derived oracle correspondences/landmarks; not available at inference. Learned-loop confidence diagnostics (held out): precision 0.2357, Brier 0.5288, ECE 0.6154. Brier/ECE are not factor-covariance calibration.
\end{table}

\begin{figure}[t]
\centering
\includegraphics[width=.5\textwidth]{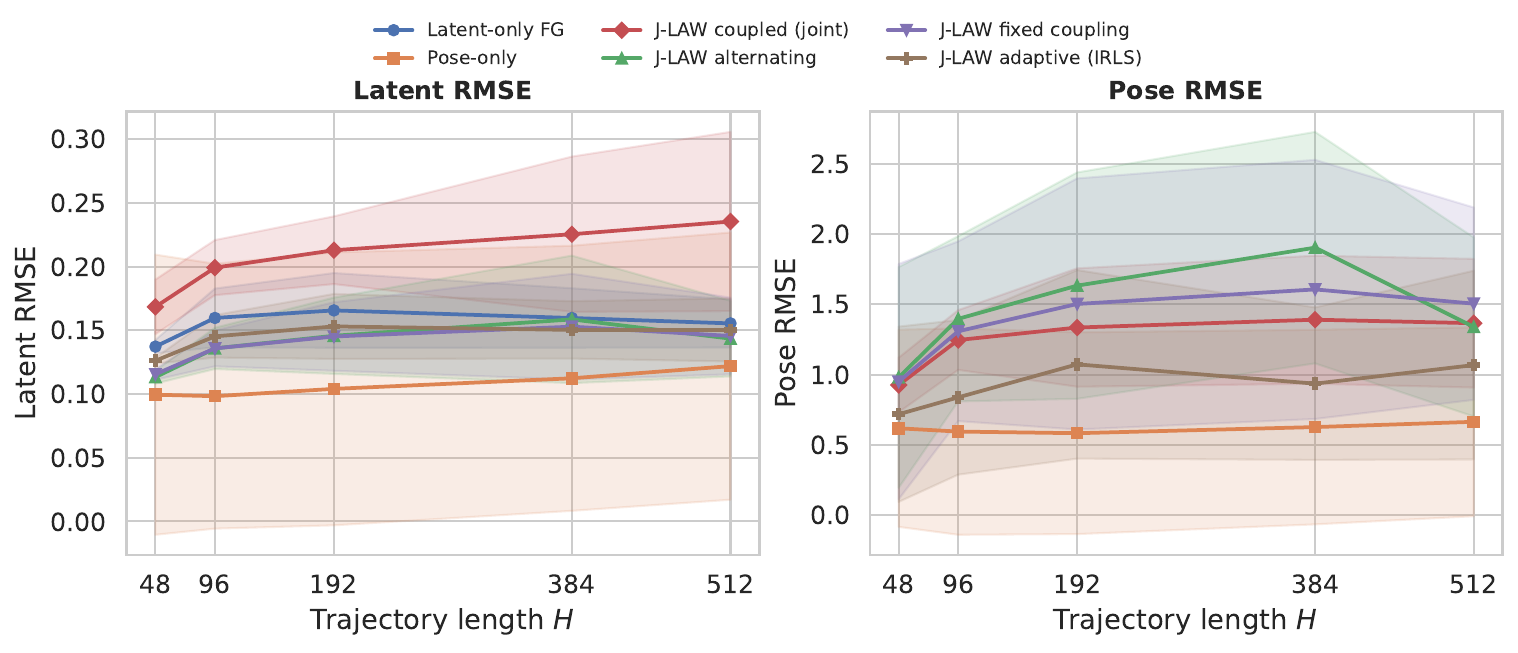}
\caption{WildGS diagnostic: latent RMSE (left) and pose RMSE (right) versus trajectory length $H$ on four held-out scenes (mean $\pm$ standard deviation). We compare direct joint optimization, alternating pose--latent inference with fixed coupling, and an IRLS-reweighted coupling variant. The trends characterize optimization and latent--pose trade-offs in this real-video diagnostic.}
\label{fig:trends}
\end{figure}

\subsection{Ablation Study on WildGS Data}

We run an ablation on 8 WildGS scenes (4 train / 4 test) using a frozen LeWM encoder ($E_\theta$) and a trained pose-to-latent coupling decoder $h_\phi$ (300 epochs, train RMSE $<$ 0.01). Figure~\ref{fig:trends} visualizes latent and pose RMSE trends across trajectory lengths. The primary modes are:
\begin{itemize}
    \item \textbf{Latent-only FG}: optimize $Z$ only (no poses, no coupling).
    \item \textbf{Pose-only}: optimize $X$ only; latents are $z_t = h_\phi(x_t)$.
    \item \textbf{J-LAW coupled}: alternating block coordinate descent over $X$ and $Z$.
\end{itemize}

To examine behavior over longer trajectories, we sweep trajectory length
$H \in \{48, 96, 192, 384, 512\}$ on the same four WildGS test scenes.
Table~\ref{tab:horizon} reports endpoint drift (latent final error) for
pose-only, open-loop, and alternating J-LAW.

\begin{table}[h!]
\centering
\small
\begin{tabular}{cccc}
\toprule
$H$ & Pose-only & J-LAW alt. & Open-loop \\
\midrule
48  & 2.25 & 2.11 & 2.43 \\
96  & 2.35 & 2.61 & 2.85 \\
192 & 2.47 & 2.31 & 2.51 \\
384 & 2.66 & 2.51 & 2.32 \\
512 & 3.11 & 2.33 & 2.33 \\
\bottomrule
\end{tabular}\\[1mm]
\caption{Endpoint latent drift versus trajectory length on four held-out
WildGS scenes. Alternating J-LAW has the lowest mean drift across the five
reported horizons (2.374), compared with pose-only decoding (2.568) and
open-loop rollout (2.488); the ordering varies by horizon.}
\label{tab:horizon}
\end{table}

%Across the five reported horizons, alternating J-LAW achieves the lowest
%mean endpoint latent drift (2.374), compared with 2.568 for pose-only
%decoding and 2.488 for open-loop rollout. J-LAW is lowest at $H{=}48$,
%$192$, and $512$, while the ordering varies at the other horizons. This
%result suggests that fusing visual latent observations with pose--latent
%coupling can improve long-window predictive-state correction over either
%baseline alone. It does not by itself establish improved metric pose
%localization.

The reliability of the learned coupling decoder is not uniform along a trajectory. We therefore replace the fixed coupling weight with an iteratively reweighted least-squares (IRLS) scheme. After each outer iteration, we compute the residual $r_t=\|h_\phi(x_t)-z_t\|_2$ and assign the coupling factor the weight $c_t=[1+(r_t/\delta)^2]^{-1}$, where $\delta$ is the median residual in the current iteration. Large residuals are therefore downweighted rather than forcing a potentially unreliable pose--latent correspondence.

\subsection{Discussion}
%Questions to anwser: Can graph correction reduce open-loop latent error; how do loop constraints affect the trade-off between consistency and pointwise latent error; and does alternating block coordinate descent behave more reliably than direct joint optimization for the coupled objective? The WildGS evaluation separates latent-only, pose-only, direct coupled, and alternating coupled variants.
Here, we address the questions raised at the beginning of the experiments.
On PushT in Table~\ref{tab:push} and WildGS
in Table~\ref{tab:wildgs-latent}, factor-graph correction substantially
reduces latent error relative to open-loop rollout. The relinearized robust
loop factors also reduce relative-loop inconsistency while preserving
pointwise latent accuracy. These results show that factor-graph correction
can reduce open-loop latent error while enforcing revisit consistency.

Under masked visual-latent anchors, learned latent-only revisits provide a
modest held-out improvement over the no-loop baseline, and persistent
latent landmarks yield a further reduction in RMSE
as shown in Table~\ref{tab:pusht-imputation}. In the separate contiguous visual outages, correctly paired proprioception
improves latent recovery compared to the result without coupling graph,
while the direct proprioception-to-latent predictor achieves slightly
lower error than coupled graph inference.
 
On WildGS horizon dependence, alternating J-LAW has the lowest average endpoint latent drift across the five reported horizons, although the ordering varies with horizon. Figure~\ref{fig:trends} further shows a latent--pose trade-off among the coupled optimization variants rather than a uniform gain in both metrics. The IRLS-reweighted variant reduces pose RMSE relative to its matched fixed-coupling alternating configuration, but it does not consistently improve latent RMSE. Notably, the pose-only baseline directly optimizes the pose-to-latent decoder against the encoded latent target. We therefore use WildGS as a real-video performance analysis of coupled inference and optimization methods, rather than evidence of a general localization improvement.

\section{Conclusion and Future Work}
%==================================================================

J-LAW introduces a unified factor-graph formulation that connects metric pose variables, action-conditioned predictive latent states, and persistent latent landmarks. On PushT data, learned revisits, persistent landmarks, and correctly paired proprioception improve offline predictive-state recovery under missing observations. On held-out WildGS video, alternating J-LAW achieves the lowest mean endpoint latent drift across the evaluated horizons, providing a diagnostic of coupled latent--pose inference. Full metric localization improvement and closed-loop planning are considered for the future work.

\bibliographystyle{IEEEtran}
\bibliography{ResearchBib.bib}

@INPROCEEDINGS{cCatal2021,
  author = {{\c{C}}atal, Ozan and Jansen, Wouter and Verbelen, Tim and Dhoedt,
	Bart and Steckel, Jan},
  title = {{LatentSLAM}: unsupervised multi-sensor representation learning for
	localization and mapping},
  booktitle = {2021 IEEE International Conference on Robotics and Automation (ICRA)},
  year = {2021},
  pages = {6739-6745},
  url = {https://doi.org/10.1109/ICRA48506.2021.9560768}
}

@MISC{Assran2025,
  author = {Mido Assran and Adrien Bardes and David Fan and Quentin Garrido and
	Russell Howes and Mojtaba and Komeili and Matthew Muckley and Ammar
	Rizvi and Claire Roberts and Koustuv Sinha and Artem Zholus and Sergio
	Arnaud and Abha Gejji and Ada Martin and Francois Robert Hogan and
	Daniel Dugas and Piotr Bojanowski and Vasil Khalidov and Patrick
	Labatut and Francisco Massa and Marc Szafraniec and Kapil Krishnakumar
	and Yong Li and Xiaodong Ma and Sarath Chandar and Franziska Meier
	and Yann LeCun and Michael Rabbat and Nicolas Ballas},
  title = {{V-JEPA} 2: Self-Supervised Video Models Enable Understanding, Prediction
	and Planning},
  year = {2025},
  archiveprefix = {arXiv},
  doi = {10.48550/arXiv.2506.09985},
  eprint = {2506.09985},
  primaryclass = {cs.AI},
  url = {https://arxiv.org/abs/2506.09985}
}

@ARTICLE{Chi2025,
  author = {Chi, Cheng and Xu, Zhenjia and Feng, Siyuan and Cousineau, Eric and
	Du, Yilun and Burchfiel, Benjamin and Tedrake, Russ and Song, Shuran},
  title = {Diffusion policy: Visuomotor policy learning via action diffusion},
  journal = {The International Journal of Robotics Research},
  year = {2025},
  volume = {44},
  pages = {1684--1704},
  number = {10-11},
  publisher = {Sage Publications Sage UK: London, England},
  url = {https://journals.sagepub.com/doi/10.1177/02783649241273668}
}

@ARTICLE{Dellaert2017,
  author = {Dellaert, Frank and Kaess, Michael},
  title = {Factor Graphs for Robot Perception},
  journal = {Foundations and Trends in Robotics},
  year = {2017},
  volume = {6},
  pages = {1-139},
  number = {1-2},
  month = {08},
  doi = {10.1561/2300000043},
  eprint = {https://www.emerald.com/ftrob/article-pdf/6/1-2/1/11046280/2300000043en.pdf},
  issn = {1935-8253},
  url = {https://doi.org/10.1561/2300000043}
}

@ARTICLE{Gornet2024,
  author = {Gornet, James and Thomson, Matt},
  title = {Automated construction of cognitive maps with visual predictive coding},
  journal = {Nature Machine Intelligence},
  year = {2024},
  volume = {6},
  pages = {820--833},
  number = {7},
  publisher = {Nature Publishing Group UK London},
  url = {http://doi.org/10.1038/s42256-024-00863-1}
}

@INPROCEEDINGS{Lee2013,
  author = {Lee, Gim Hee and Fraundorfer, Friedrich and Pollefeys, Marc},
  title = {Robust pose-graph loop-closures with expectation-maximization},
  booktitle = {2013 IEEE/RSJ International Conference on Intelligent Robots and
	Systems},
  year = {2013},
  pages = {556-563},
  url = {https://doi.org/10.1109/IROS.2013.6696406}
}

@MISC{Maes2026,
  author = {Lucas Maes and Quentin Le Lidec and Damien Scieur and Yann LeCun
	and Randall Balestriero},
  title = {{LeWorldModel}: Stable End-to-End Joint-Embedding Predictive Architecture
	from Pixels},
  year = {2026},
  archiveprefix = {arXiv},
  doi = {10.48550/arXiv.2603.19312},
  eprint = {2603.19312},
  primaryclass = {cs.LG},
  url = {https://arxiv.org/abs/2603.19312}
}

@MISC{Mur-Labadia2026,
  author = {Lorenzo Mur-Labadia and Matthew Muckley and Amir Bar and Mido Assran
	and Koustuv Sinha and Mike Rabbat and Yann LeCun and Nicolas Ballas
	and Adrien Bardes},
  title = {{V-JEPA} 2.1: Unlocking Dense Features in Video Self-Supervised Learning},
  year = {2026},
  archiveprefix = {arXiv},
  doi = {10.48550/arXiv.2603.14482},
  eprint = {2603.14482},
  primaryclass = {cs.CV},
  url = {https://arxiv.org/abs/2603.14482}
}

@MISC{Rao2026,
  author = {Pratyaksh Rao and Wancong Zhang and Randall Balestriero and Yann
	LeCun and Giuseppe Loianno},
  title = {{SkyJEPA}: Learning Long-Horizon World Models for Zero-Shot Sim-to-Real
	Control of Quadrotors},
  year = {2026},
  archiveprefix = {arXiv},
  eprint = {2606.23444},
  primaryclass = {cs.RO},
  url = {https://arxiv.org/abs/2606.23444}
}

@INBOOK{Thrun2008,
  pages = {13--41},
  title = {Simultaneous Localization and Mapping},
  publisher = {Springer Berlin Heidelberg},
  year = {2008},
  editor = {Jefferies, Margaret E. and Yeap, Wai-Kiang},
  author = {Thrun, Sebastian},
  address = {Berlin, Heidelberg},
  booktitle = {Robotics and Cognitive Approaches to Spatial Mapping},
  doi = {10.1007/978-3-540-75388-9_3},
  isbn = {978-3-540-75388-9},
  url = {https://doi.org/10.1007/978-3-540-75388-9_3}
}

@INPROCEEDINGS{Zheng2025,
  author = {Zheng, Jianhao and Zhu, Zihan and Bieri, Valentin and Pollefeys,
	Marc and Peng, Songyou and Armeni, Iro},
  title = {{WildGS-SLAM}: Monocular Gaussian Splatting SLAM in Dynamic Environments},
  booktitle = {2025 IEEE/CVF Conference on Computer Vision and Pattern Recognition
	(CVPR)},
  year = {2025},
  pages = {11461-11471},
  url = {https://doi.org/10.1109/CVPR52734.2025.01070}
}
\end{document}